\documentclass[conference]{IEEEtran}
\IEEEoverridecommandlockouts
\usepackage{cite}
\usepackage{amsmath,amssymb,amsfonts}
\usepackage{algorithm}
\usepackage{graphicx}
\usepackage{textcomp}
\usepackage{algpseudocode}
\usepackage{setspace}
\usepackage{subcaption}
\usepackage{caption}
\usepackage{graphics} 
\usepackage{epsfig} 
\usepackage{epstopdf}
\usepackage{booktabs}
\usepackage{subcaption}
\usepackage{gensymb}
\usepackage{multirow}
\usepackage{float}
\AppendGraphicsExtensions{.tif}

\def\BibTeX{{\rm B\kern-.05em{\sc i\kern-.025em b}\kern-.08em
    T\kern-.1667em\lower.7ex\hbox{E}\kern-.125emX}}
\begin{document}

\title{ 
A Deep Reinforcement Learning Approach for Dynamically Stable Inverse Kinematics of Humanoid Robots\\
{\footnotesize }
\thanks{*Equal contribution}
}

\author{\IEEEauthorblockN{Phaniteja S\textsuperscript{*} }
\IEEEauthorblockA{\textit{Robotics Research Center} \\
\textit{IIIT Hyderabad}\\
Hyderabad, India \\
phaniteja.sp@gmail.com}\\
\IEEEauthorblockN{Abhishek Sarkar}
\IEEEauthorblockA{\textit{Robotics Research Center} \\
\textit{IIIT Hyderabad}\\
Hyderabad, India \\
abhishek.sarkar@iiit.ac.in}
\and
\IEEEauthorblockN{Parijat Dewangan\textsuperscript{*} }
\IEEEauthorblockA{\textit{Robotics Research Center} \\
\textit{IIIT Hyderabad}\\
Hyderabad, India \\
parijat.dewangan@research.iiit.ac.in}
\and
\IEEEauthorblockN{Pooja Guhan}
\IEEEauthorblockA{\textit{Robotics Research Center} \\
\textit{IIIT Hyderabad}\\
Hyderabad, India \\
pooja.guhan@students.iiit.ac.in}\\
\IEEEauthorblockN{K Madhava Krishna}
\IEEEauthorblockA{\textit{Robotics Research Center} \\
\textit{IIIT Hyderabad}\\
Hyderabad, India \\
mkrishna@iiit.ac.in}
}

\maketitle

\begin{abstract}
Real time calculation of inverse kinematics (IK) with dynamically stable configuration is of high necessity in humanoid robots as they are highly susceptible to lose balance. This paper proposes a methodology to generate joint-space trajectories of stable configurations for solving inverse kinematics using Deep Reinforcement Learning (RL). Our approach is based on the idea of exploring the entire configuration space of the robot and learning the best possible solutions using Deep Deterministic Policy Gradient (DDPG). The proposed strategy was evaluated on the highly articulated upper body of a humanoid model with 27 degree of freedom (DoF). The trained model was able to solve inverse kinematics for the end effectors with 90\% accuracy while maintaining the balance in double support phase.

\end{abstract}


\section{Introduction}
\begin{figure}
\center
\includegraphics[width=0.8\columnwidth,scale=0.15]{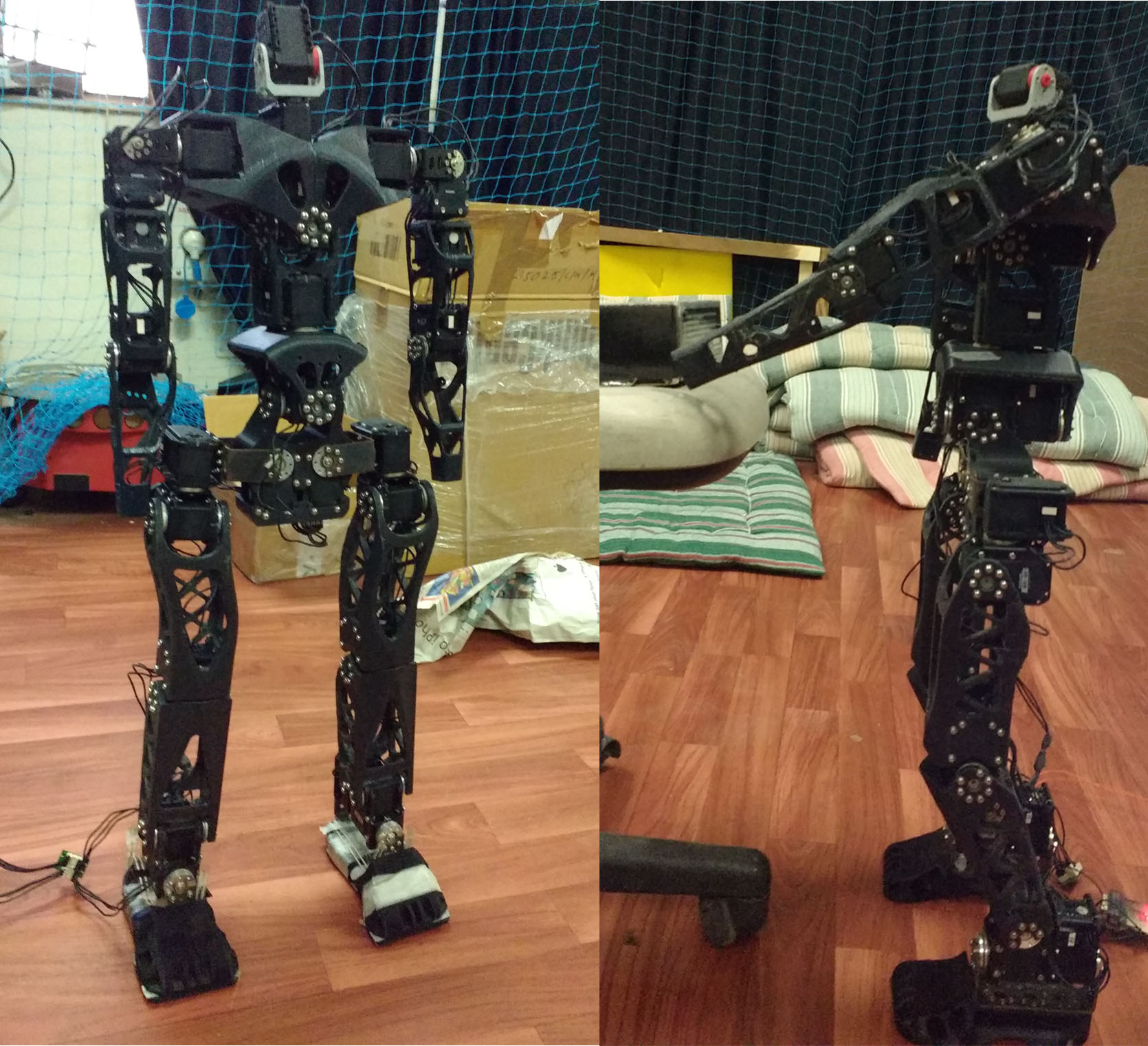}
\caption{Humanoid Robot with articulated torso}
\label{bot}
\end{figure}
In robotic systems, the tasks are usually defined in coordinate space, whereas the control commands are defined in actuator space. In order to perform task level robot learning, an appropriate transformation from coordinate space to actuator space is required. If the intrinsic coordinates of a manipulator are defined as a vector of joint angles $\boldsymbol{\theta} \in \bf{R}^n$, and the position and orientation vector of the end effector as a vector $\bf{x} \in \bf{R}^m$, then the forward kinematics function can be given by the following equation
\begin{equation}\label{fk}
\bf{x} = \it{f}(\boldsymbol{\theta})
\end{equation}
The inverse kinematics problem\cite{ik,robust} is to find a mapping from the end-effector coordinates to actuator space which can be represented as 
\begin{equation}
\boldsymbol{\theta} = \it{f^{-1}}(\bf{x})
\end{equation}
For redundant robotic systems, that is, when the dimension of the task-space is smaller than the dimension of the joint-space $(n > m)$, $f^{-1}(.)$ is not a unique mapping. Given a task-space position $\bf{x}$, there can be many corresponding joint-space configurations of $\boldsymbol{\theta}$. Thus, learning inverse kinematics relates to learning multi-valued function.

Inverse kinematics for humanoid robots are important for applications like pick and place\cite{pickplace}, physics engines\cite{mujoco,vrep,gazebo,usarsim,bullet} and human-robot interactions like tele-operating a robot to grasp objects\cite{telop}, or execute a series of coordinated gestures\cite{imitation,im2}. Inverse kinematic approaches can be broadly divided into two categories, namely closed-from analytical methods and numerical methods. Some examples of numerical methods are BFGS\cite{bfgs}, pseudo-Jacobian inverse\cite{ik,jp,jacob}, Jacobian transpose\cite{ik,jt2,jacob}, Damped Least Square method (DLS)\cite{ml,jacob}, and Cyclic Coordinate Descent (CCD). Unlike closed-form analytical methods, the convergence time of numerical methods may vary and the results are not repeatable. On the top of that, computing inverse kinematics under constraints of stability and self-collision avoidance cannot be done efficiently in real time.

Recent advancements in RL\cite{sutton} like Deep $Q$-learning (DQN)\cite{dqn}, Deterministic Policy gradients (DPG)\cite{dpg}, Guided Policy Search\cite{gps}, Trust region policy optimization\cite{trpo} and DDPG\cite{ddpg,deeprl} provide us many frameworks for not only learning the complex problem of IK, but also to optimize the required criteria. Among these methods, DDPG learns an efficient policy when the action space is continuous which is the case with inverse kinematics. Reinforcement Learning works on the experienced data, and thus would avoid problems due to matrix inversions which may occur while solving general inverse kinematics. Therefore, learning would never demand impossible postures which occurs due to ill-conditioned matrix inversions. 



In this paper we demonstrate how deep RL can be used to learn generalized solutions for inverse kinematics. We propose a DDPG based IK solver which takes into account criteria of stability and self-collision avoidance while generating configurations. We validated the method by applying this framework to learn reachability tasks in the double support phase\cite{dsp,dekker2009zero}.

The rest of the paper is organized as follows. In Section II, kinematic model of the robot is explained followed by the calculation of zero moment point (ZMP) and a brief description of general IK solvers. Section III explains DDPG algorithm and the proposed methodology to learn the stable IK solver. It also gives a brief description about the reward function used and the network architecture. Following that we show the results of training in Section IV along with numerical simulations. Finally conclusions and future work are discussed in Section V.

\section{Kinematic modelling and Inverse Kinematics}
The humanoid model used for study is shown in Fig. \ref{bot}. The robot is small, with total height of 84 cm, total weight below 5 Kg and 27 DoF. The main highlight is its vertebral column (5 DoF articulated torso), which makes the biped more close to human. In this work, we have tried to explore the usage of articulated torso for performing reachability hand tasks.

\subsection{Kinematic model of the robot}
\begin{figure}[ht]
\center
\includegraphics[width=0.8\columnwidth]{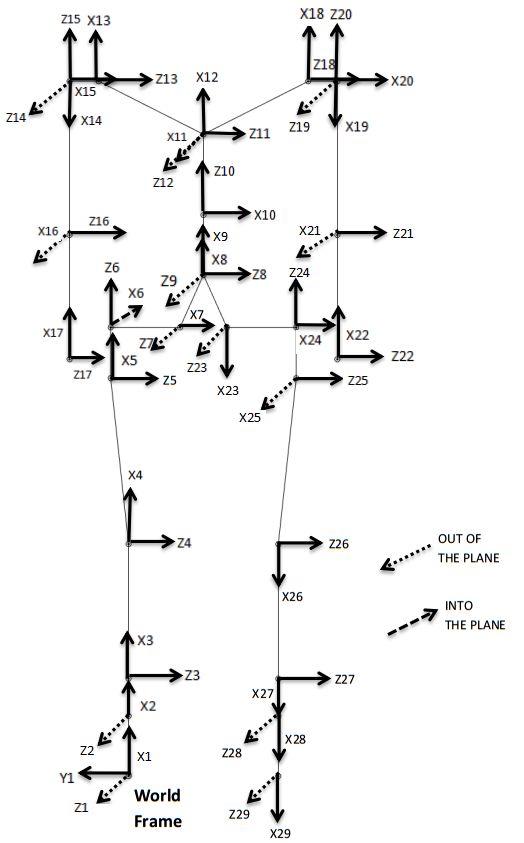}
\caption{Kinematic model of the robot}
\label{kinmodel}
\end{figure}

The kinematic model of the robot is represented using the D-H convention with the base frame at the right leg sole and the first joint angle starting from right ankle. Starting from the base, a coordinate frame is defined at each joint and at the end of each end-effector (hands and left leg). The complete kinematic model with axes numbering is shown in Fig. \ref{kinmodel}.

In Fig. \ref{kinmodel}, all frames are right handed and hence  only X and Z axes are shown for the frames in order to have a simpler representation. Y axes can be easily identified by using right hand thumb rule. The world frame is located at the right foot sole and is oriented as shown in Fig. \ref{kinmodel}.

\subsection{Stability and Calculation of ZMP}
ZMP\cite{zmp}, \cite{dekker2009zero} is one of the widely used dynamic stability measures which was proposed by Vukobratovi\'c and Stepanenko in 1972. In bipeds, a configuration is stable if ZMP lies inside the convex hull of the feet, which defines the support polygon\cite{supportpolygon}.

In double support phase when the robot is stationary, ZMP is equal to the center of mass (CoM) projection in the support polygon.  Whereas when the robot is not stationary, ZMP might deviate from the CoM projection. In order to have an accurate ZMP point we need to include momentum and angular momentum into the calculation. Hence the ZMP should be calculated as \cite{kajita2014introduction}

\begin{equation}\label{ZMP}
\begin{split}
p_x = \frac{ Mgx - \dot{L_y}}{Mg+\dot{P_z}}\\p_y = \frac{ Mgy + \dot{L_x}}{Mg+\dot{P_z}}
\end{split}
\end{equation}

where $x$, $y$ are the x and y coordinates of CoM, M is the total Mass of the robot, $g$ is acceleration due to gravity and [$L_x$, $L_y$, $L_z$], [$P_x$, $P_y$, $P_z$] are the angular and linear momentum respectively w.r.t base frame. 

\subsection{General Inverse Kinematics and Stability}
Most of the general inverse kinematics solvers work in the velocity domain and solve for inverse kinematics using Jacobian or gradient descent method. 
Although constraints like singularity avoidance and joint limits can be included in these methods, stability criteria cannot be included directly in IK solver. Hence the resultant solution of inverse kinematics may not be stable in case of humanoids.

One way to avoid such scenarios is to keep on checking the intermediate configurations and reiterating until a stable configuration is achieved by the IK solver. Generally, this takes very long time to converge and in some cases, the solver might not be able to find a stable configuration at all. Therefore, there is a need for an IK solver which takes stability into account while solving and doesn't require any external checking. This kind of solver can be easily learnt using learning based inverse kinematics\cite{d2001learning}. RL requires only the kinematic model of the robot for learning a generalized solver. Using this idea, a generalized IK framework can be defined for complex robots like humanoids where balance and posture plays a great role apart from reaching the goal.

\section{Deep Deterministic Policy Gradient (DDPG) for Inverse Kinematics}
Reinforcement learning \cite{sutton} can be used to train an agent which learns from the environment directly without the use of any external data. An RL agent takes an action ($a$) depending on its state ($s$) and observes the reward ($r$) given out by the environment. This process is repeated. The aim of the RL agent is to maximize the cumulative reward in any task. Hence the underlying reward function and its modelling plays a crucial role in RL. 

A RL agent can learn different types of policies pertaining to the given task. A policy function represents the agent's behaviour needed to complete a given task. It is a mapping from state to action. Policy learning can be subdivided into two categories: 1) Stochastic 2) Deterministic. Stochastic policy learns the conditional probability of taking an action $a$, given that it is in a state $s$, $\pi(a/s) = P[A_t = a/ S_t =s]$. A stochastic policy is useful only when number of actions are discrete and countable at any given state. In a continuous state-action space, this leads to a large number of possibilities and hence large memory and search time. A deterministic policy on the other hand learns the action as a function of state, $a = \pi(s)$. This function is non-linear in complex systems like humanoids and neural-networks serves as good model to learn such kind of functions. DDPG provides us with a very good frame work to train the neural-networks for learning highly non-linear functions.

\subsection{Deep Deterministic Policy Gradient}
DDPG uses the underlying idea of DQN in the continuous state-action space. It is an Actor-Critic Policy learning\cite{ac} method with added target networks to stabilize the learning process. DDPG uses experience replay which addresses the issue of data being dependent and non-identically distributed as most optimization algorithms need samples that are identical and independently distributed. Transitions are sampled from the environment according to the given exploration policy and the tuple $(s_t,a_t,r_t,s_{t+1})$ are stored in a replay buffer of finite size. When this buffer becomes full, oldest samples are discarded. A mini-batch of samples, $m_b$ is used to update the network. The critic network $Q(s,a,w)$ is learnt using Bellman equation\cite{bellman} as in $Q$-learning\cite{qlearn}, and the actor updates the policy in the direction that improves $Q$, i.e., critic provides the loss function for actor. In order to avoid the divergence of neural networks in $Q$-learning,  target networks are used which track the original networks slowly. 

Suppose $Q(s,a,w)$, $\mu(s,\theta)$ represent critic and actor networks respectively and $Q'(s',a,w')$, $\mu'(s',\theta')$ represent their target networks, the loss function for critic network can be given as

\begin{equation}\label{Lc}
L_c = (r + {\gamma Q(s',\mu(s', \theta'), w')} - Q(s,a,w))^2
\end{equation}
where $r$ is the reward and $\gamma$ is discount factor.

The loss function for the actor is given as
\begin{equation}\label{La}
L_a = \nabla_{a} Q(s,a,w) \nabla_{\theta}\mu(s,\theta)
\end{equation}

The target networks are updated as follows with $\tau << 1$
\begin{equation}\label{tn}
w' = \tau w + (1-\tau)w' \qquad
\theta' = \tau \theta + (1-\tau)\theta'
\end{equation}

As it can be observed from Eq. \ref{Lc}, reward function is an integral part of the network update and hence the underlying policy that is learnt by the network. Therefore reward function should be modelled carefully so that the RL agent learns the policy correctly.

\subsection{State vector and network architecture} 
The chosen state vector consists of joint angles ($\bf{q}$), the end effector coordinates and the goal position coordinates. The action vector is a set of angular velocities, $\dot{\bf{q}}$. Hence the policy learns a mapping from configuration space to velocity space. The state vector is of 21 dimensions and the action vector is of 13 dimensions. A 2 layered network consisting of fully connected layers with 400, 300 hidden units is used for both Actor and Critic. $cRelu$\cite{crelu} is taken as activation function and $\tau$ is taken as 0.001. Batch normalization is used in the network to avoid over-fitting and handle the scale variance problems.

\subsection{Reward function}
The main objective of an IK problem is to provide a set of angles ($\bf{q}$) that are needed to reach the given position and orientation. Most of the Jacobian based methods solve for this using gradient descent and the solution is minimized in terms of $\dot{\bf{q}}$. Therefore $min(\dot{\bf{q}})$ is included as a part of the reward function. In order to ensure that the configurations given out by the solver are within the stability region, a large negative reward is given whenever it goes out of stability bounds. The final reward function is shown in Eq. \ref{r_f}.

\begin{equation}\label{r_f}
 r =
\begin{cases}
 - \alpha dist - \beta \sqrt{\sum_{i} ({\Delta q_i})^2} & if \, stable \, {and} \, collision free \\
 -\kappa & if \, unstable \\
\end{cases}
\end{equation}
where $\alpha , \beta, \kappa$ are the normalization constants, $dist$ is the absolute distance between goal position and the current end effector position and $\Delta q_i$ is the angular difference between the starting configuration and the current configuration of the $i_{th}$ joint. In our case, $\alpha$ is $\frac{1}{70}$, $\beta$ is $\frac{10}{2\pi}$ and $\kappa$ is 20.

\begin{algorithm}
\caption{Humanoid Environment}\label{env}
\begin{algorithmic}[1]
\Statex {{\bf function}{$\ Reset()$}}
\Statex $\quad config \gets Set\ random\ initial\ configuration$
\Statex $\quad goal\ \gets Set\ random\ goal\ position$
\Statex $\quad s\ \gets GetState(config)$
\Statex {\bf return}{$\ s$}
\Statex 
\Statex {{\bf function}{$\ GetState(config)$}}
\Statex $\quad EnfPos \gets ForwKin(config)$
\Statex $\quad state \gets concat(config,EnfPos,goal,done)$
\Statex {\bf return}{$\ state$}
\Statex 
\Statex {{\bf function}{$\ Step(action)$}}
\Statex $\quad action \gets clip(action,ActionBound)$
\Statex $\quad config \gets config + action$
\Statex $\quad ForwKin(config)$ \Comment{Updates the kinematic model}
\Statex $\quad r, done \gets  Reward(config)$
\Statex $\quad s=GetState(config)$
\Statex {\bf return}{$\ s, r, done$}
\Statex 
\Statex {{\bf function}{$\ Reward(config)$}}
\Statex $\quad ZMP \gets CalZMP(config)$
\Statex $\quad ${{\bf if }{$ZMP\ in\ support\ polygon\ {\bf{and}} \ collision\ free\ $}{\bf then}}
\Statex $\qquad  r = - \alpha dist - \beta \sqrt{\sum_{i} ({\Delta q_i})^2}$
\Statex $\quad${\bf else}
\Statex $\qquad  r = -\kappa$
\Statex $\quad${\bf end if}
\Statex $\quad${{\bf if}{$\ goal\ is\ reached\ $}{\bf then}}
\Statex	$\qquad r = r + \lambda$ \Comment{Add large positive reward}
\Statex $\qquad done = True$
\Statex {$\quad${\bf else}}
\Statex $\qquad done = False$
\Statex $\quad${\bf end if}
\Statex {\bf return}{$\ r,done$}
\end{algorithmic}
\end{algorithm}
\subsection{Environment modelling and Training}
Modelling of the environment is a very crucial part for any RL algorithm. In order to learn a generalized inverse kinematics solution, the entire configuration space needs to be spanned while training. This is achieved by randomly sampling both start configuration and goal position for every episode. Algorithm \ref{env} shows the environment used for the training.

The Actor and Critic networks are trained using the given Humanoid Environment. In DDPG, policy is learnt by the Actor network and Q-value function is learnt by the Critic network. Target networks are used for both Actor and Critic and these are updated very slowly using $\tau$ as in Eq. \ref{tn}. We used a Replay buffer of size $5 \times 10^5$. The pseudo code for training is given in Algorithm \ref{ddpg}. A normally distributed decaying random noise is used for the exploration noise which is observed to provide good results in training. Critic and Actor networks are updated as given in Eqs. \ref{Lc} and \ref{La} respectively. The training results and their evaluations are shown in the subsequent sections.
\begin{algorithm}
\caption{IK learning using DDPG}\label{ddpg} 
\begin{algorithmic}[1]
\State $Randomly\ initialize\ Actor\ and\ Critic\ Networks$
\State $TargetActorNet \gets ActorNet$
\State $TargetCriticNet \gets CriticNet$
\For{$i = 1\ to \ MaxEpisodes$}
\State $s \gets Reset()$
\For{$j = 1\ to\ MaxStep$}
\State $action \gets Policy(s)$ \Comment{Get action using ActorNet}
\State $action \gets action + N$ \Comment{Add Exploration Noise}
\State $s',r,done \gets Step(action)$
\State $ReplayBuffer \gets Store(s,a,s',r)$
\If{$size(ReplayBuffer) > BSize$}
\State $batch \gets RandSample(ReplayBuffer,BSize)$
\State $Q \gets Update(CriticNet,batch)$ 
\Statex $\qquad \qquad \qquad \triangleright $ Q-value function update
\State $Policy \gets Update(ActorNet,batch,Q)$
\Statex $\qquad \qquad \qquad \triangleright $ Policy update
\State $Update\ Target\ networks\ using\ \tau$
\EndIf
\If{$done$}
\State $break$
\EndIf
\EndFor
\EndFor
\end{algorithmic}
\end{algorithm}

\begin{figure*}[t]
        \centering
        \begin{subfigure}[h]{0.3\textwidth}
			\includegraphics[width=\textwidth]{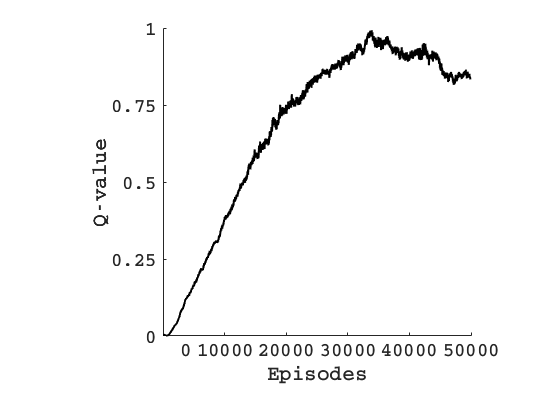}
			\caption{Normalized Q-value}
			\label{qval}
			\end{subfigure}
        \begin{subfigure}[h]{0.3\textwidth}
				\includegraphics[width=\textwidth]{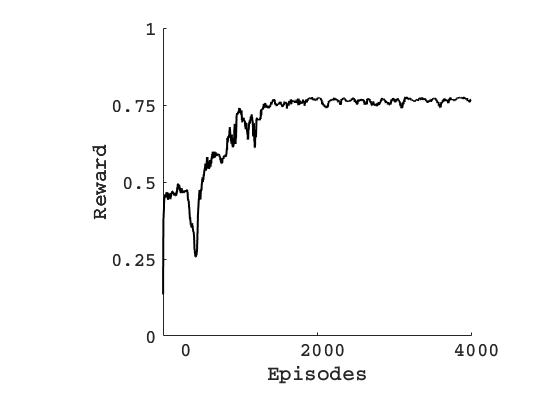}
                \caption{Normalized Reward}
				\label{rew}
        \end{subfigure}
        \begin{subfigure}[h]{0.3\textwidth}
				\includegraphics[width=\textwidth]{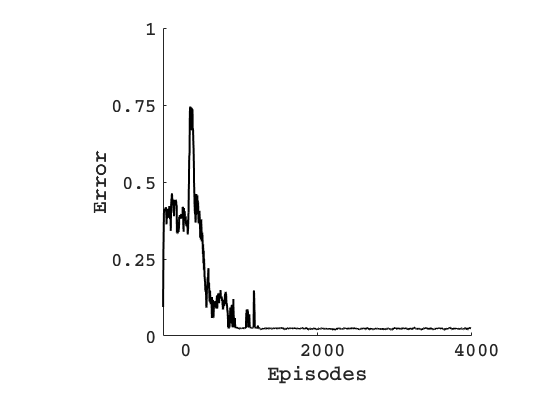}
				\caption{Error}
				\label{dist}
        \end{subfigure}
        \caption{Results of Training on 7.5 million steps}
        \label{result}
\end{figure*}
\begin{table*}[t]
  \centering
  \caption{Performance of the IK Solver}
  \resizebox{0.8\textwidth}{!}{
    \begin{tabular}{c|ccccccccc}
    \toprule
    \multirow{1}[4]{*}{Accuracy} & & &  & & Episodes&  &  \\
               & 900  & 1200  & 1500 & 1800 & 2100  & 2700 & 3000 & 3900 & 9900\\
    \midrule
    Min      & 23\% & 61\%   & 72\%   & 75\% & 79\%  & 80\% & 76\% & 88\% & 87\%\\
    Max     & 36\% & 69\%   & 78\%   & 82\% & 85\% & 84\% & 83\% & 89\% & 93\%\\
    Mean     & 30\% & 66.33\%  & 75\%  & 78.33\% & 82.66\% & 82\% & 79.66\% & 88.66\% & 90\%\\
    
    \bottomrule
    \end{tabular}%
    }
  \label{accu}
\end{table*}%
\section{Results and Simulations}
The humanoid model was trained taking into account all the criteria explained in the previous sections. Training was run for 50000 episodes with 150 steps in each episode, totalling 7.5 million steps.

\subsection{Training results}\label{AA}
Figs. \ref{qval} and \ref{rew} show the normalized $Q$-value and reward of training. In Fig. \ref{qval}, the plot started to nearly saturate after 30000 episodes showing the attainment of optimal $Q$-value function. Error is defined as the difference between the end-effector and goal position at the end of an episode. The corresponding normalized error plot is shown in Fig. \ref{dist}. The error goes on decreasing with training and reaches a minimum value soon after 1500 episodes showing that the network has learnt the required policy. The same is reflected in the reward function as shown in Fig. \ref{rew}.

The trained model is tested for reachability tasks by giving random start configurations and goal positions. The accuracies of the network obtained by using 100 random samples over 3 different seeds at different points of learning are documented in Table. \ref{accu}.
It can be observed from the table, that the accuracy goes on increasing with the training and oscillates in a small region after 2000 episodes. The highest mean accuracy obtained is 90\% at 9900 episodes. 
\subsection{Simulated Experiments}
The trained IK solver is tested in the dynamic simulator of MSC Adams environment. The joint trajectories generated by the solver are given as input to the simulator for testing the solution. A set of three experiments which have high probability of losing balance are chosen in order to demonstrate the efficiency of the learnt IK solver and also to explore the advantages of an articulated torso. 
\begin{figure}[h]
        \centering
        \begin{center}\includegraphics[width=8cm,height=6cm,keepaspectratio]{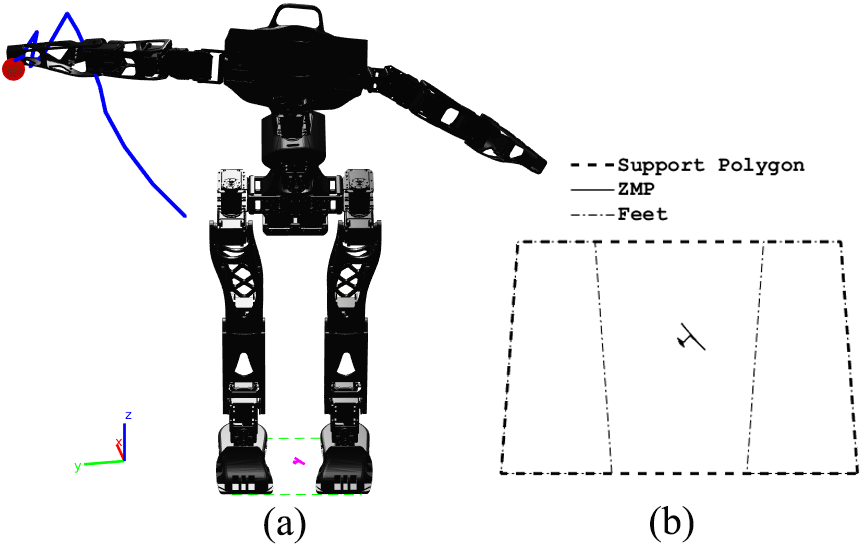}
				\caption{Trajectory and ZMP plot for Task 1}
        \label{task1}
        \end{center}
\end{figure}

\begin{figure}
        \begin{center}
\includegraphics[width=9cm,height=6cm,keepaspectratio]{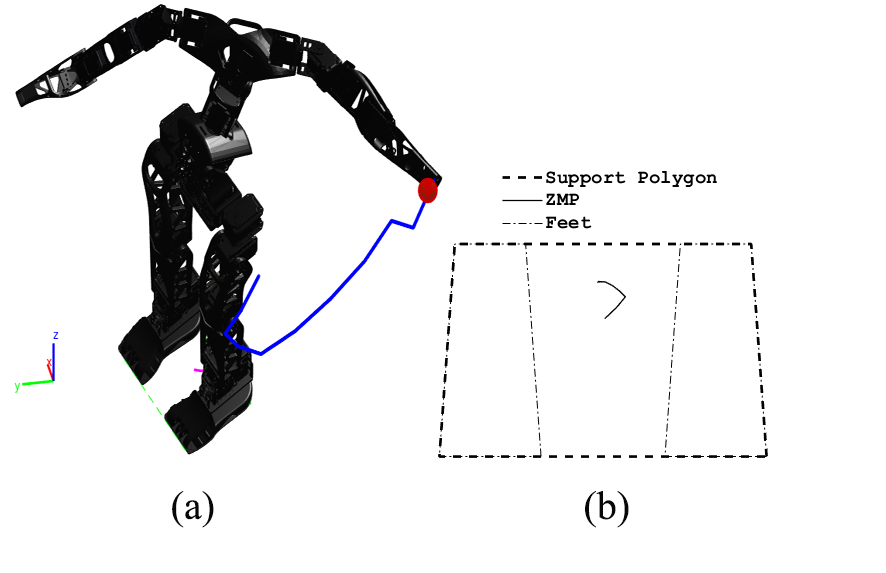}
                \caption{Trajectory and ZMP plot for Task 2}
        \label{task2}
        \end{center}
\end{figure}

\begin{figure}
        \begin{center}
\includegraphics[width=9cm,height=6cm,keepaspectratio]{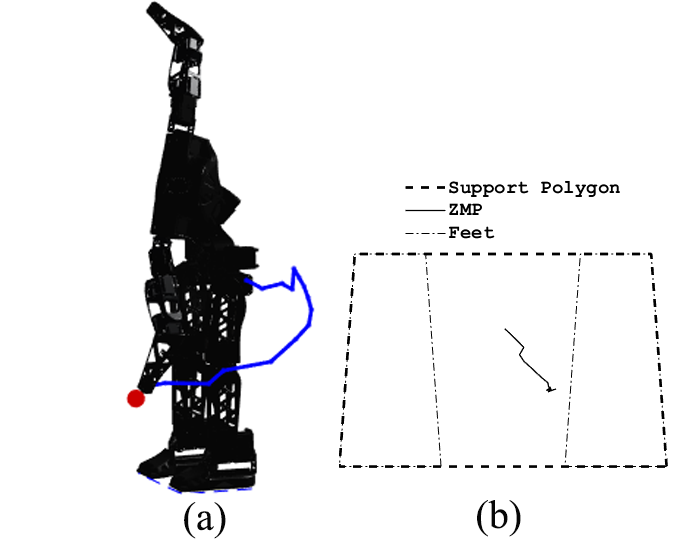}
				\caption{Trajectory and ZMP plot for Task 3}
                \label{task3}
        \end{center}
\end{figure}
In the first task, it had to reach a point in the far right end where it needs to use its spine to bend towards the right, as shown in Fig. \ref{task1}a. In the second task, as shown in Fig. \ref{task2}a, it has to reach a point in the left-back side, where the chest motion is tested. In the last task, it had to reach a point below its knee where it tried to explore the limitation of the pelvis and abdomen joints which is shown in Fig. \ref{task3}a. 

Figs. \ref{task1}a, \ref{task2}a and \ref{task3}a show the end effector trajectories along with the final posture of the robot. The corresponding ZMP plots for tasks are shown in Figs. \ref{task1}b, \ref{task2}b and \ref{task3}b. It was observed that the ZMP stays within the support polygon while performing each of these tasks.

In all of the three tasks, it was observed that the robot used the other hand to balance itself and stay within the stability region and also avoided self collision. The vertebral column played an important role in making the postures similar to humans, which can be observed from the Figs. \ref{task1}, \ref{task2} and \ref{task3}.



\section{Conclusions and Future work}
This paper proposes a methodology for generating dynamically stable inverse kinematic solutions using deep RL. The approach was able to learn a robust IK solver within 2000 episodes. The robustness of the model was tested by giving various complex tasks. It was able to reach most of the points in its configuration space without losing its balance. Also, the solver converges to an inverse kinematics solution in less number of iterations as compared to general inverse kinematic solvers in most of the cases. Although the proposed model has limitations on precision, this model can serve as a good prototype for inverse kinematics solver of highly redundant manipulators.   

Future work includes increasing the accuracy of the trained model and learning more complex tasks which involves movement of legs.
\bibliographystyle{IEEEtran}
\bibliography{reference}

\end{document}